\documentclass[conference]{IEEEtran}
\IEEEoverridecommandlockouts
\usepackage{cite}
\usepackage{amsmath,amssymb,amsfonts}
\usepackage{algorithmic}
\usepackage{algorithm}
\usepackage{graphicx}
\usepackage{textcomp}
\usepackage{xcolor}
\usepackage[normalem]{ulem}
\def\BibTeX{{\rm B\kern-.05em{\sc i\kern-.025em b}\kern-.08em
    T\kern-.1667em\lower.7ex\hbox{E}\kern-.125emX}}
\begin{document}

\title{Client and Training Data Selection for Computationally Efficient Synchronized Federated Learning
\thanks{%
Accepted for publication in the 29th Euromicro Conference
on Digital System Design (DSD 2026).
Author's accepted manuscript.
\par
\copyright{} 2026 IEEE. Personal use of this material is
permitted. Permission from IEEE must be obtained for all
other uses, in any current or future media, including
reprinting/republishing this material for advertising or
promotional purposes, creating new collective works,
for resale or redistribution to servers or lists,
or reuse of any copyrighted component of this work
in other works.
}
}

\author{\IEEEauthorblockN{Muzaffer Citir$^\dagger{}$, Hiroki Nishikawa$^\dagger{}^\dagger$, Sangyoung Park$^\dagger{}$}
\IEEEauthorblockA{
$^\dagger{}${Smart Mobility Systems}, \textit{Technical University of Berlin, Germany} \\ 
$^\dagger{}^\dagger${Graduate School of Information Science and Technology}, \textit{The University of Osaka, Japan} \\
$^\dagger{}$\{muzaffer.citir, sangyoung.park\}@tu-berlin.de,
$^\dagger{}^\dagger$nishikawa.hiroki@ist.osaka-u.ac.jp}
}


\maketitle

\begin{abstract}
Federated learning (FL) is a promising paradigm of machine learning, which preserves user privacy by enabling learning without sharing raw data with a cloud server.
Straggling clients have been a problem for FL as they introduce delays in aggregating the local models and hence, the convergence of the global model.
Therefore, it is important to have a mechanism that ensures fast convergence of the global model as well as good FL participation rate.
Another issue for the convergence of a model in FL is the non-independent and identically distributed (non-iid) data across the clients.
Prior approaches based on probabilistic client selection do not work well under non-iid data especially when the number of clients is small.
We show scenarios where such approaches fail and propose a joint client-training data selection algorithm for fast convergence of FL models.
Our experiments on CIFAR-100 dataset show that convergence of the FL model can be significantly improved over prior works that can consider non-iid data and heterogeneous computation and higher model accuracy.
\end{abstract}


\section{Introduction}
Federated learning (FL) has emerged as a paradigm, enabling a multitude of edge devices to engage in distributed learning while upholding user privacy and data security \cite{mcmahan2017communication}. 
The concept involves edge devices training local ML models on their own data and periodically sharing model parameters with each other, thus maintaining data privacy \cite{bonawitz2019towards, lim2020federated, gu2021server}.
Despite its advantages in data privacy, parallel processing, and reducing data transmission delays to cloud servers (CS), FL still faces challenges in time-sensitive ML applications~\cite{xiao2023time}.
First, model training is performed on heterogeneous edge devices where computing and communication resources are limited.
Therefore, overall model training could take an indefinitely long time if model aggregation is delayed due to a few stragglers~\cite{cui2022helcfl, reisizadeh2022straggler, zhao2021federated}.
Second, data is often non-independent and identically  distributed (non-iid) across the edge devices.
In case of classification problems, the model accuracy could be undermined if certain classes are not represented well in the training dataset because they are located in resource-constrained devices, and therefore, cannot participate in FL well.

\begin{figure*}[t!]
    \centering
    \includegraphics[width=\textwidth]{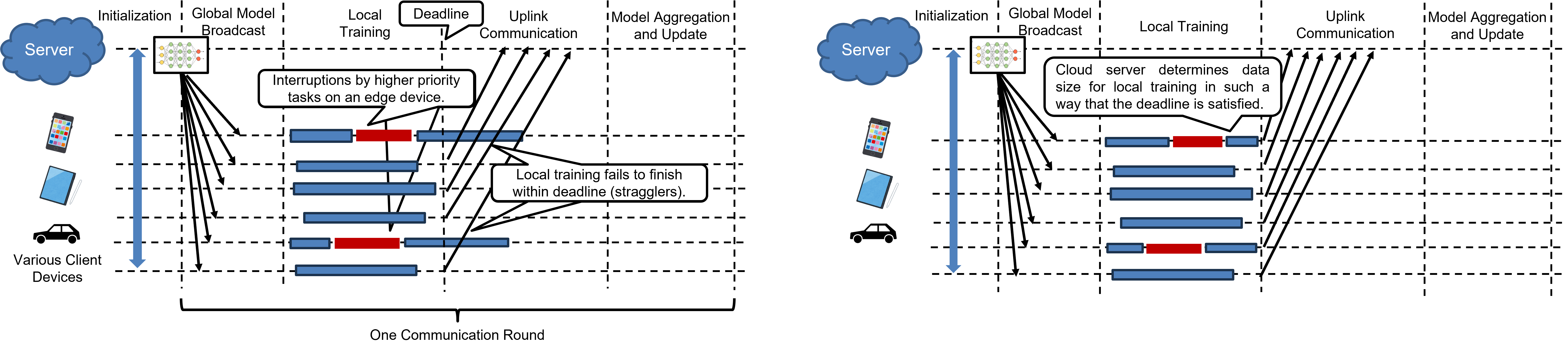}
    \vspace{-5mm}
    \caption{(a) A deadline miss example of FL devices scheduling, and (b) avoiding it with proper data allocation.}
    \label{fig:example}
\end{figure*}

Fig.~\ref{fig:example} depicts how stragglers could impact the performance of FL training.
Here, we assume synchronous update of the model where a CS waits for all participants in the FL round before aggregating the model.
Local training on edge devices is generally regarded as a best-effort service, which can be interrupted by higher-priority tasks. This results in straggling edge devices, worsening the performance of model training and slowing the convergence of the FL model~\cite{cui2022helcfl, gao2021fedswap}.
We also assume a deadline (or timeout) by which local training must finish, as the CS cannot indefinitely wait for stragglers in an FL iteration.
Fig.~\ref{fig:example}(a) shows how a lack of computing capacity and interruptions can lead to deadline misses in such an FL setting.
In contrast, if an FL system is aware of the computation capacity and has stochastic information about potential causes of delay such as interruptions, it could ask the edge device to train on fewer data samples to ensure completion before the deadline (Fig.~\ref{fig:example}(b)).

Apart from meeting the FL timing constraints, which is mainly concerned with computing capacity and the amount of training data, the quality of training data is also crucial for the accuracy of the global model.
Although prior works proposed client selection and resource management schemes for FL under a deadline~\cite{yu2021jointly,li2019smartpc}, most of them assume iid data, where the convergence of the model performance is less tricky~\cite{ma2022state}.
Even if edge devices are able to complete the training and send the model to the CS in time, the convergence of the global model will be slow unless the training dataset is sufficiently balanced.
To address such issues, this paper proposes a device selection and data allocation algorithm for edge FL that minimizes the overall model training time under a timing constraint per FL iteration and non-iid data across the edge devices.
It is assumed that the CS cannot predict exactly when the training on participating devices finishes due to the uncertainty in training and interruptions by other tasks, the algorithm takes a probabilistic approach and tries to maximize the expected value of the training data that can contribute to model aggregation each FL round.
The algorithm also tries to achieve a balanced class distribution during the client selection and when it is determining the training dataset size.
The contributions of this paper are as follows:
\begin{itemize}
    \item We show the impact of the training dataset size and class distribution on the training times and model accuracy
    \item We incorporate a probabilistic approach for estimating training times and task interruptions
    \item A joint device and non-iid-aware training data selection algorithm for determining the training data size and balance between classes.
\end{itemize}

The rest of this paper is organized as follows: Section~\ref{sec:related_work} summarizes the related work. Section~\ref{sec:system_model} describes the probabilistic system model backed by measurements. Section~\ref{sec:algorithm} proposes the joint client and data selection algorithm for fast FL model convergence. Section~\ref{sec:experiments} shows the experimental results and a comparison with the state of the art. Finally, Section~\ref{sec:conclusions} concludes this paper with future remarks.

\section{Related Work}\label{sec:related_work}


Numerous studies on FL looked into optimization of communication and computational resources to boost system efficacy. Wang et al. introduced algorithms to enhance resource utilization and diminish energy expenditure in 5G edge networks by modulating the frequency of global aggregation at edge servers~\cite{wang2019adaptive}.
Broadening this scope, Wadu et al. examined client scheduling and resource allocation, introducing a strategy that employs Gaussian process regression for channel prediction to mitigate accuracy loss in FL~\cite{wadu2020federated}. This strategy was subsequently enhanced to incorporate the computational resources of clients, presenting a more integrated perspective on resource optimization~\cite{wadu2021joint}. In a similar fashion, Li et al. devised SmartPC, a system with dual-level controllers designed to optimize the FL learning process, showing the potential efficiency gains from device selection and configuration adjustments~\cite{li2019smartpc}.
These works, however, consider iid data across the clients, which means the data quality does not change significantly depending on client selection. 

In a related research direction, Amiri et al. tackled the optimization of communication resources through device scheduling policies that consider both channel conditions and the significance of local model updates under non-iid data \cite{amiri2020update}. This work was further elaborated with a convergence analysis and a model update compression strategy \cite{amiri2021convergence}. Furthermore, to address latency issues, Xia et al. and Xu et al. proposed client scheduling methodologies grounded in multi-armed bandit theory, aimed at reducing training latency—a crucial aspect that highlights the complex nature of efficiency in FL systems \cite{xia2020multi,xu2021online}.
A heuristic method to select computationally efficient (powerful hardware) and statistically efficient (useful data)~\cite{lai:usenix21} has been proposed based on the loss value after local training. Such methods use metrics, which are available only after the local training, therefore wasting precious computation resources even though the client is not selected in that FL round.

Wang et al. later introduced Fed-LBAP and FedMinAvg methods designed to regulate workload distribution to minimize computational time and accuracy loss during FL model training \cite{wang2020optimize}. They also presented an algorithm, MinCost, to tackle the optimization challenge \cite{wang2020towards}, and proposed OLAR \cite{pilla2021optimal}, an optimal greedy solution to reduce computational time in FL training. Furthering this line of research, subsequent work developed scheduling algorithms for FL aimed at reducing energy consumption by managing workload distribution \cite{pilla2023scheduling}, predicated on the notion that training's computational latency can be modulated by workload allocation \cite{shi2020device}.
An approach that modifies the client selection probability based on the gradient value has proven to be effective~\cite{chen:iot24}.
Unlike these approaches, in our method, clients that do not participate in FL do not waste any computation resources, and that most \textit{helpful} clients are deterministically selected.

Our work is distinguished from other works in that i) it considers non-iid data and explicitly balances class distribution for an FL round, ii) it does not require local training to complete for client selection, and iii) deterministically selects the most helpful clients.
To the best of the authors' knowledge, no other works have simultaneously considered all these points.
The details and effectiveness of the proposed approach are elaborated in the rest of the paper.

\section{System Model}\label{sec:system_model}

\subsection{Overall FL Architecture}
The FL system consists of a single Cloud Server (CS) and $M$ edge devices, denoted as $M = \{1, 2, \ldots, M\}$.
Each device $i$ possesses a local dataset $D_i = \{x_{i,n} \in \mathbb{R}^s, y_{i,n} \in \mathbb{R}\}_{n=1}^{|D_i|}$. 
This paper assumes that each device $i$ uses a part of $D_i$ for training, denoted as $d_i$, to meet the deadline of an iteration, $T^{dead}$.
Here, $x_{i,n}$ denotes $n$-th $s$-dimensional input data vector at device $i$, and $y_{i,n}$ the corresponding labeled output for $x_{i,n}$.
Here we also make similar assumption to existing FL papers such that the amount of dataset size as well as the class information is known to the device~\cite{mcmahan2017communication}.

The primary objective of FL training is to determine the model parameter $w$ that minimizes a loss function to the entire dataset, which is given by:
\begin{equation}
    \min_w \,  F(w) := \min_w \frac{1}{M} \sum_{i\in M} f_i(w),
\end{equation}
where the local loss function $f_i(w)$ on the dataset $d_i$ is defined as $f_i(w) = \frac{1}{|d_i|} \sum_{n\in d_i} f(w, x_{i,n}, y_{i,n})$, and $f(w, x_{i,n}, y_{i,n})$ captures the error of the model parameter $w$ on the input-output pair $\{x_{i,n}, y_{i,n}\}$.

Our FL uses an iterative approach to solve Equation (1). Each round, indexed by $k$, contains the following three steps: (1) The CS broadcasts a global model $w_k$ to a number of edge devices, which are assumed to be under randomly selection (RS) by the CS and participate in the round. (2) Each device $i$ updates its local model by a gradient descent algorithm on the local dataset (i.e., $w_{k+1}^i = w_k - \eta \nabla f_i(w_k)$, where $\eta$ is a learning rate), and uploads the updated model $w_{k+1}^i$ to the CS until a deadline. If a device misses the deadline, the updated model is disposed. (3) The CS aggregates all the models uploaded until the deadline to generate a new global model $w_{k+1} = \frac{1}{|\Pi_k|} \sum_{i\in\Pi_k} w_{k+1}^i$.

\subsection{Probabilistic Modeling of Time-sensitive FL}

Hereafter, we consider an arbitrary round $k$ in this paper and omit the index without loss of generality. We model the total latency as the sum of three components: computation latency ($t_{\text{cp}}$), interruption latency ($t_{\text{intr}}$), and communication latency ($t_{\text{comm}}$). The cumulative probability that the total latency satisfies a given deadline, $T_{\text{dead}}$, is expressed as:
\begin{equation}
P(t_{\text{cp}} + t_{\text{intr}} + t_{\text{comm}} \leq T_{\text{dead}}).
\label{eq:total_latency}
\end{equation}
The CS determines the training dataset size $d_i$ for each edge device to improve the learning rate while ensuring this probabilistic constraint is satisfied. Below, we detail the modeling of each latency component.

\textbf{Computational Latency: } The computation latency $t_{\text{cp}}$ of local training on an edge device, given a dataset size $d_i$, is modeled using a shifted exponential distribution:
\begin{equation}
P(t_{\text{cp}} \leq t) = 1 - \exp\left(-\frac{\mu_i}{d_i} (t - a_i d_i)\right),
\label{eq:computation_latency}
\end{equation}
where $a_i > 0$ represents the device's maximum computation capacity, and $\mu_i > 0$ indicates the fluctuation in computation capacity. The minimum computation time is given by $a_i d_i$, and the probability increases as $t$ grows. The parameters $a_i$ and $\mu_i$ are determined by the performance of the edge device and the characteristics of the FL application.
The mean and variance of $t_{\text{cp}}$ are:
\begin{equation}
\mathbb{E}[t_{\text{cp}}] = a_i d_i + \frac{d_i}{\mu_i}, \quad \text{Var}[t_{\text{cp}}] = \frac{d_i^2}{\mu_i^2}.
\label{eq:computation_latency_stats}
\end{equation}

\begin{figure}[t]
\centering
\includegraphics[width=0.45\textwidth]{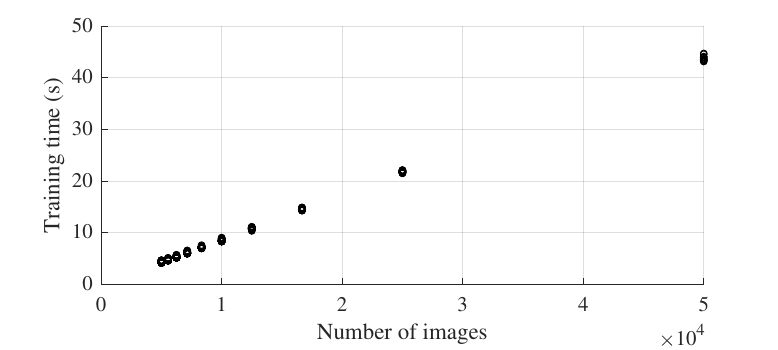}
\caption{Training time vs dataset size for CIFAR-10 dataset.}
\label{fig:time_vs_data}
\end{figure}
\begin{figure}[t]
\centering
\includegraphics[width=0.45\textwidth]{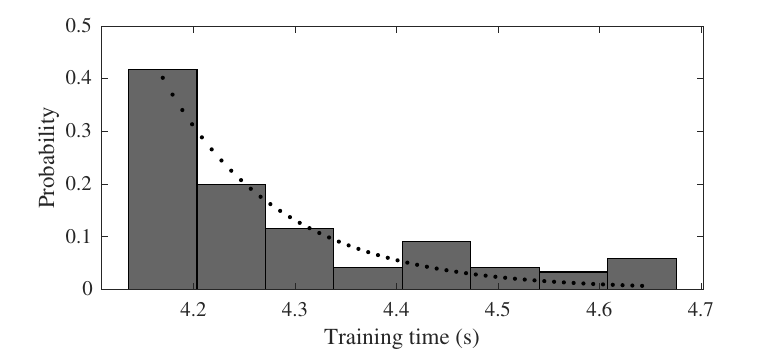}
\caption{Time distribution with 5$\times$$10^3$ images of CIFAR-10.}
\label{fig:histogram}
\end{figure}

In order to validate the presented computation latency model, we have conducted preliminary experiments that measure the training times of CIFAR-10 dataset~\cite{krizhevsky2009learning} using the hardware specified in Table~\ref{tab:hardware_specs}.
In the validation, we have run 660 times of training with CIFAR-10 on the environment and measured the training times.
Fig.~\ref{fig:time_vs_data} shows the result, where x- and y-axes show the number of images used to training and the measured training time (secs), respectively.
This measurement indicates that we have observed the linear relationship between the training dataset size and training time.
Next, we depict a histogram with 120 runs to yield the probability distribution of training times, which roughly follows the exponential distribution, as shown in Fig. \ref{fig:histogram}.
An example $\mu_i$ and $a_i$ values for our setup as a result of curve fitting is also shown in the graph.
In conclusion, it has been observed that the results of the model are well aligned with the outcomes of empirical experiments although it cannot be overlooked that the selected model may not perfectly mimic computation delay in every aspect.
Throughout this paper, we assume that the CS is aware of the computation capabilities of edge devices, i.e., different versions of smartphones, infrastructure monitoring sensors, etc., and the application characteristics that dictate $\mu_i$ and $a_i$.

\begin{table}[t]
\centering
\caption{Hardware specifications for FL Training}
\label{tab:hardware_specs}
\footnotesize
\begin{tabular}{|l|l|}
\hline
\textbf{Component}       & \textbf{Specification}           \\ \hline \hline
CPU                      & Intel i9-12900KF 3,2 GHz        \\ 
GPU                      & NVIDIA GeForce RTX 4090 24 GB (CUDA 12.2)   \\ 
RAM                      & 64 GB DDR5-SDRAM 4800 MHz        \\ 
Operating System         & Ubuntu 22.04.3 LTS                 \\ \hline \hline
\end{tabular}
\end{table}

\textbf{Interruption Latency: } The interruption latency $t_{\text{intr}}$ arises due to preemption by other tasks on edge devices, such as a smartphone user launching an application \cite{kim:iccad15}. This is modeled using an $M/M/1$ queuing system:
\begin{equation}
P(t_{\text{intr}} \leq t) = 1 - \exp(-\lambda_{\text{intr}} t),
\label{eq:interruption_latency}
\end{equation}
where $\lambda_{\text{intr}}$ represents the rate of interruptions (Poisson process), and the service rate follows an exponential distribution with parameter $\mu_{\text{intr}} \geq \lambda_{\text{intr}}$, where $\mu_{\text{intr}}$ represents the service rate of interruptions, i.e., how quickly interruption tasks are handled by the edge device.
The mean and variance of $t_{\text{intr}}$ are:
\begin{equation}
\mathbb{E}[t_{\text{intr}}] = \frac{1}{\mu_{\text{intr}} - \lambda_{\text{intr}}}, \quad \text{Var}[t_{\text{intr}}] = \frac{1}{(\mu_{\text{intr}} - \lambda_{\text{intr}})^2}
\label{eq:interruption_latency_stats}
\end{equation}

\textbf{Communication Latency: } Communication latency $t_{\text{comm}}$ occurs when devices upload locally trained parameters to the CS. We model this latency as a Gaussian distribution:
\begin{equation}
P(t_{\text{comm}} \leq t) = \Phi\left(\frac{t - \mu_{\text{comm}}}{\sigma_{\text{comm}}}\right),
\label{eq:communication_latency}
\end{equation}
where $\mu_{\text{comm}}$ and $\sigma_{\text{comm}}$ are the mean and standard deviation of the communication latency, respectively.
The mean and variance of $t_{\text{comm}}$ are:
\begin{equation}
\mathbb{E}[t_{\text{comm}}] = \mu_{\text{comm}}, \quad \text{Var}[t_{\text{comm}}] = \sigma_{\text{comm}}^2.
\label{eq:communication_latency_stats}
\end{equation}

The weak dependency between training latency and communication latency allows us to approximate them as independent; thus assuming independence among the three components for simplicity, the cumulative probability can be expressed as:
\begin{equation}
\begin{split}
P(t_{\text{total}} \leq T_{\text{dead}}) = & \, P(t_{\text{cp}} \leq T_{\text{dead}}) \cdot \\
& P(t_{\text{intr}} \leq T_{\text{dead}}) \cdot P(t_{\text{comm}} \leq T_{\text{dead}})
\end{split}
\label{eq:independent_prob}
\end{equation}

\subsection{Problem Formulation}
To ensure the deadline constraint is satisfied, the CS optimizes the dataset size $d_i$ for each device. The optimization problem is formulated as:
\begin{equation}
\max_{d_i} \sum_{i} U_i(d_i),
\label{eq:objective_function}
\end{equation}
subject to:
\begin{equation}
U_i(d_i) = P(t_{\text{total}} \leq T_{\text{dead}}) \cdot d_i ,
\end{equation}
\begin{equation}
P(t_{\text{total}} \leq T_{\text{dead}}) \geq 1 - \epsilon,
\end{equation}
where $U_i(d_i)$ represents the utility of training that is expressed as the expected value of the probability of the total latency with dataset size $d_i$, $\alpha$ is a parameter to consider the trade-off between the accuracy and latency, and $\epsilon$ is the acceptable probability of exceeding the deadline that depends on the requirement for a real-time application.

Furthermore, we proceed under the assumption that latency to aggregate models on the CS is negligible thanks to its stronger computation capability.

\section{Joint Client and Data Selection Algorithm}\label{sec:algorithm}

Before describing the joint client and data selection algorithm in detail, we make the following assumptions supporting the logic of the algorithm.

\noindent\textbf{Assumption 1:} A larger training datasize contributes to faster convergence of the FL model accuracy.

\noindent\textbf{Assumption 2:} Using unseen data contributes to higher model accuracy over FL iterations.
If a few clients are repeatedly selected due to higher computing capacity, their local data might be re-used for training. Many prior works do not have to consider this as the participating clients are probabilistically selected every FL round. As the proposed algorithm deterministically selects clients, it should avoid re-using the same data simply to maximize the training dataset size.

\noindent\textbf{Assumption 3:} Impact of unbalanced collective class distribution significantly reduces the test accuracy of under-represented classes.

\begin{algorithm}[t!]
 \small
 \caption{ Joint client and data selection algorithm}
 \begin{algorithmic}[1]
 \renewcommand{\algorithmicrequire}{\textbf{Input:}}
 \renewcommand{\algorithmicensure}{\textbf{Output:}}
 \REQUIRE A set of clients $K$, disjoint local dataset $D_k=D_{k,1}\cup D_{k,2}\cup \cdots \cup D_{k,m}$ where $D_{k,i}$ is the set of images of class $i$ in client $k$, computational capability $a_k$, fluctuation $\mu_k$, and interruption parameters $\lambda_k$ and $\mu_{intr,k}$ and average number of times an image has been used for training $n_{avg,k}$ per client.
 \ENSURE Subset of clients $l\in K$, training dataset size per class $|TD_{k,i}|\forall k\in l,\forall i$
 \FORALL{$k\in K$}
    \STATE $|TD_k|\leftarrow \textrm{argmax}_{|TD|}(|TD|\cdot P_k(t^{cp}+t^{itr}+t^{cm}\leq T^{dead}))$
    \STATE $|TD_{k,i}|\forall i\leftarrow \textrm{assignEven}(D_k, |TD_k|)$
 \ENDFOR
\STATE $u_k\leftarrow w_1 \cdot (1-n_{avg,k})|TD_k|+w_2\cdot \textrm{min}(|TD_{k,1}|,|TD_{k,2}|,\cdots),$\\$\forall k \in K$
 \STATE $l\leftarrow \phi$
 \STATE $D_{cur}\leftarrow \phi$
  \WHILE{$|l|<n_{target}$}
    \STATE $k\leftarrow \textrm{max}_{u_k}(K)$
    \STATE $l\leftarrow l\cup k$
    \STATE $\displaystyle n_{avg,k}\leftarrow n_{avg,k} + \frac{|TD_k|}{|D_k|}$
    \STATE $TD_{cur}\leftarrow TD_{cur}\cup TD_k$
    \STATE $K\leftarrow K-k$
    \STATE $i\leftarrow \textrm{argmin}_i(|TD_{cur,i}|)$
    \STATE $cc_m\leftarrow \text{classCoverage($TD_{cur},m$)}\forall m$
    \STATE $u_m\leftarrow w_1\cdot exp(-n_{avg,m}/10)\cdot|TD_m|+w_2 \cdot |D_{k,i}|+w_3 \cdot cc_m, \forall m\in K$
 \ENDWHILE
 
 \RETURN $l$, $|TD_{k,i}|\forall k\in l$,$\forall$ i, $n_{avg,k}\forall k \in K$
 \end{algorithmic} 
 \label{alg:algorithm}
\end{algorithm}

Based on the assumptions, we design the as described in
Algorithm~\ref{alg:algorithm}.
We assume that the algorithm is running on the CS and is aware of the set of all clients $K$, sizes of local datasets and their class distribution on each client $|D_{k,i}|$ where $k\in K$ and $i$ is the class index, parameters characterizing the computing capacity $a_k$ and $\mu_k$, interruption parameters $\lambda_k$ and $\mu_{intr,k}$, and $n_{avg,k}$ denoting the average number of times an image has been used for training over the FL rounds.
The output of the algorithm is $l\subset K$ a set of clients selected for this FL iteration, $|TD_{k,i}|$ number of images per class in the selected clients, and updated $n_{avg,k}$ for the next FL iteration.

In line 2, the algorithm determines the size of the training dataset, $|TD_k|$ that maximizes the expected dataset size for all client $k$, i.e., dataset size multiplied by the probability the training finishes within the deadline. In line 3, training dataset sizes per class $|TD_{k,i}|$ are derived by assignEven() function.
The function is a heuristic for determining $|TD_{k,i}|$ for all $i$ such that the distribution is as even as possible while the sum is equal to $|TD_{k}|=\sum_i |TD_{k,i}|$. If there is enough data from all classes, $|TD_{k,i}|=|TD_k|/N_{classes}$.
If the number of images for certain classes is smaller than $|TD_k|/N_{classes}$, then all images from the classes are used for training and the rest of the classes are assigned the same number of images such that the training dataset size is equal to $|TD_k|$.

Lines 8 to 16 describe the essence of the algorithm, where clients are greedily selected based on the usefulness metric $u_k$, which is defined as follows.
\begin{equation}\label{eq:usefulness}
    u_k\leftarrow w_1\cdot exp(-n_{avg,k}/10)|TD_k|+w_2 \cdot |D_{k,i}|+w_3\cdot cc_k, \forall k\in K,\\
\end{equation}
where $i=\textrm{argmin}_i(|TD_{cur,i}|)$ is the index of the class that has the least amount of training data allocated so far, $|TD_{cur}|$, in this FL round.
The usefulness metric of a client $k$ is a weighted sum of three elements, where the first is the training dataset size multiplied by a decreasing exponential function of the average number of times an image has been used in that client.
The reasoning behind this is that the convergence of the FL model is faster when a larger training dataset is used (assumption 1), but still want to prevent using the data likely from high computing capacity devices over and over again. Eventually, $u_k$ forces FL to select other clients through the term $n_{avg,k}$.
The second term is used to give preference to clients, which have images belong to the class that are  in this FL round. This effectively balances the class distribution of the training dataset to be used in this FL round (assumption 2).
The third term is a supplementary term ensuring the class coverage at the warm up phase of the loop (lines 8-16) before the second term starts to effectively work.

Until the target number of clients is reached, the client with largest $u_k$ is selected in the while loop.
The client $k$ is added to the set $l$ (line 10), $n_{avg,k}$ is updated (line 11) for the next FL round, and $TD_{cur}$ is updated with the newly selected client's data (line 12), $k$ is removed from candidate client set $K$ in this FL round (line 13), the class with the fewest amount of images is identified (line 14), and finally, the usefulness metric is updated for every client still in $K$.
Before the while loop at line 8, $TD_{cur}$ is empty, and the class with the least number of images cannot be determined.
Therefore, the second term of $u_k$ is instead initialized with the number of images from the class that has the fewest samples in each respective client (line 5).

The outputs $l$, the client selection, $|TD_{k,i}|\forall k\in l, \forall i$, the training dataset are derived, and $n_{avg,k}$ will be used in subsequent FL iterations.
Actual training dataset is selected randomly, i.e., random selection of $|TD_{k,i}|$ images out of $D_{k,i}$. 

\section{Experiments}\label{sec:experiments}

In this section, we present the experimental results and compare with existing works.
The two baselines we consider are \textit{MinCost}~\cite{wang2020towards} and \textit{probPart} based on~\cite{chen:iot24}, which can deal with non-iid data and computation heterogeneity.
The algorithms in the work do not assume a fixed deadline like our work, but aim at minimizing the time required for global model convergence.
Regarding MinCost algorithm, the sum of the training dataset across clients should be $D$, which is an input to the algorithm.
The algorithm uses parameter $\alpha$ to address the non-iidness when assigning data.
Higher value discourages clients with fewer number of classes to participate in the FL rounds.
We used the value of 2.5 as it is in the middle of the range the work has suggested.
As for probPart algorithm, the usefulness of a client under non-iid data is characterized in the $G_i$ parameter.
The values of $G_i$ for client $i$ have been calculated using a training on a small batch of the training data. 
Our version of probPart is simplified that selection probability is set just to be proportional to $G_i$, but preserves the essence of the algorithm.

Here, we evaluate the performance of our proposed FL scheduling algorithm on the CIFAR-100 dataset~\cite{krizhevsky2009learning}, which is used for evaluation and consists of 50,000 training and 10,000 test images for the experiments.
In local training, we employ MobileNetV2~\cite{sandler2018mobilenetv2}, a lightweight CNN architecture with depthwise separable convolutions, inverted residual blocks, batch normalization, and \textit{ReLU6} activation functions. 
This configuration results in a total of 3.5M parameters.
The training utilizes the AdamW optimizer with weight decay and cross-entropy loss function.
Additionally, we configure the federated system to operate without data compression.

\subsection{Performance of Data Allocation Algorithm}

In this subsection, we present the convergence patterns of the aforementioned FL schemes.
We assumed a hardware setup where five types of devices are participating in FL each having different computation capacities.
For example, clients are assumed to have on average the computation capability of completing the training for 500 data points in 15 seconds with an 85\% probability, where a coefficient of variation (CV) is 75\%.

\begin{figure}[t]%
\centering
\includegraphics[width=\columnwidth]{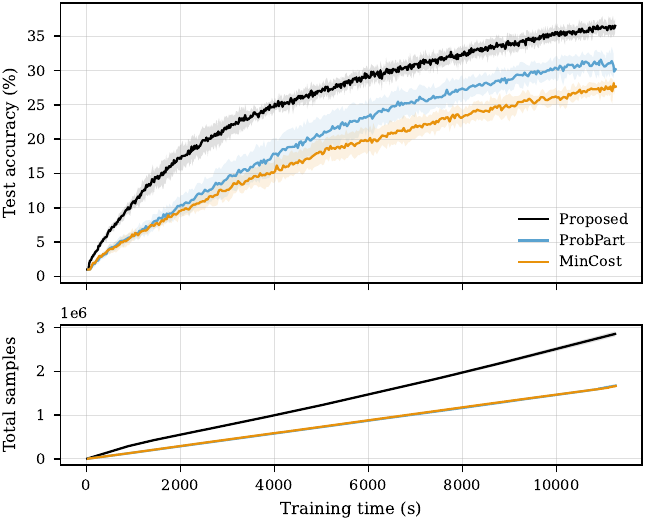}
\caption{Test accuracy over time for the proposed, probPart and MinCost algorithms for CIFAR-100 dataset distributed across 50 clients. Shaded area denotes 1 std range over 6 experiments with random seeds for data distribution.}
\vskip 0pt
\label{fig:final_result}
\end{figure}

\begin{figure*}[t]
\centering
\includegraphics[width=\textwidth]{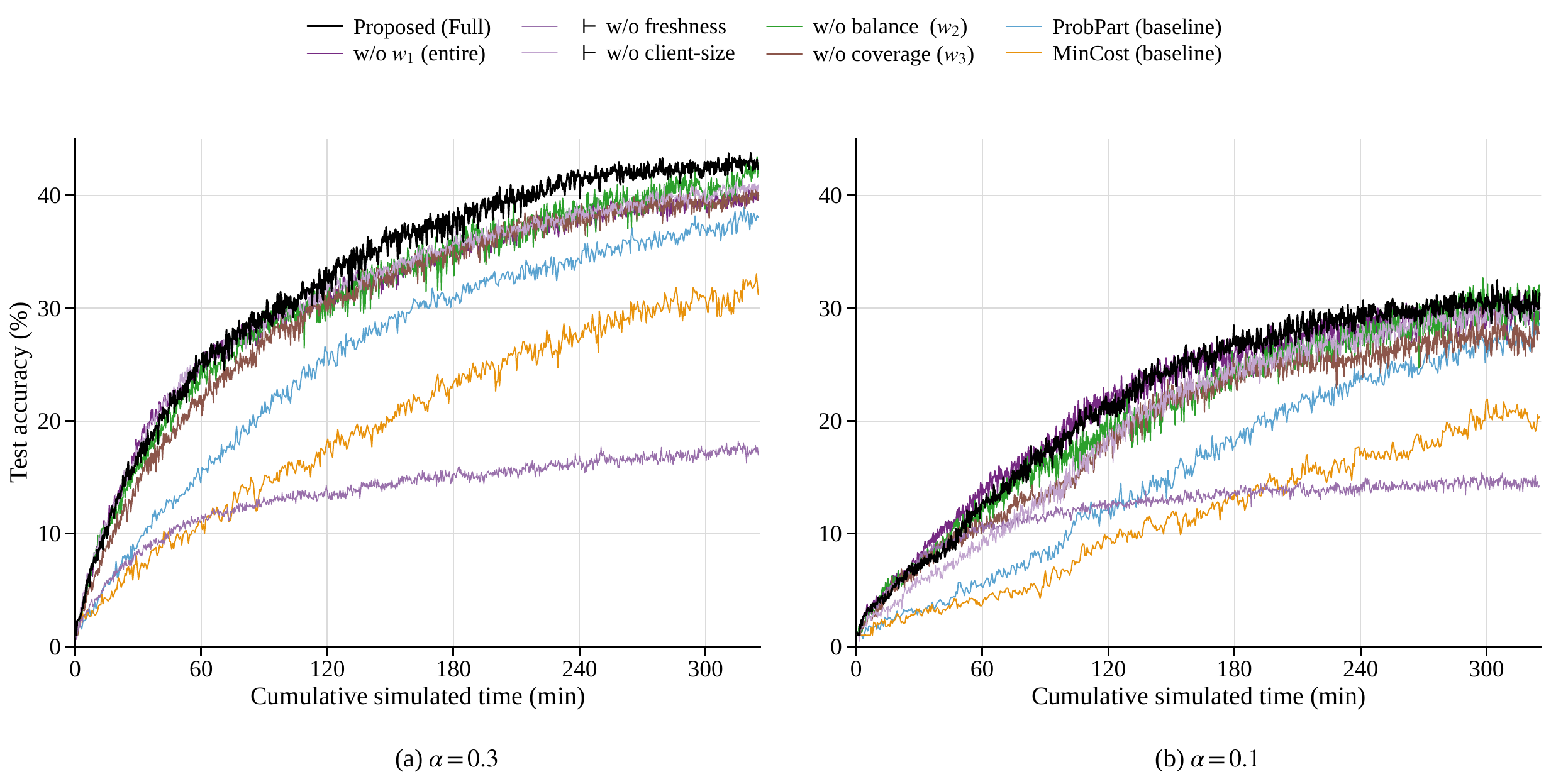}
\caption{Component ablation of the usefulness metric (Eq.~\ref{eq:usefulness}) as test accuracy over training time, under two degrees of non-iid severity:
(a) $\alpha=0.3$ and (b) $\alpha=0.1$. Each variant disables one term of $u_k$; for $w_1$, its two sub-factors (the freshness penalty and the client dataset-size preference) are disabled separately. \textit{Proposed} keeps all terms, while probPart and MinCost are shown for reference. The experiments use the same client, model, and deadline configuration as Fig.~\ref{fig:final_result}, and are reported for a single random seed.}
\label{fig:ablation}
\end{figure*}

\noindent\textbf{Model convergence speed and test accuracy:}
Fig.~\ref{fig:final_result} shows the experimental results for the three FL scenarios where each client is assumed to have disjoint pre-allocated 1000 images.
In order to generate non-iid data across clients, i.e., classes being unevenly distributed, we use label-skewed Dirichlet partitioning scheme often used in similar studies~\cite{hsu2019measuring}.
The parameter determining the degree of non-iidness, $\alpha$, is set to 0.3.
We assume a total of 50 clients and for each FL round, the algorithm selects 10 clients.
The number of local epochs inside an FL iteration is set to be 5.
Deadline per FL iteration for the proposed algorithm is set to be 15 seconds.
One FL iteration for the baseline algorithms may take longer than 15 seconds or 15 seconds depending on which clients are selected.

The two baseline algorithms are synchronous FL, but not deadline-based like the proposed algorithm, i.e., they wait for the slowest device to complete the training before model aggregation using FedAvg.
The two baseline models require pre-defined training dataset sizes, which is set to 500.
To make the comparison fair among the algorithms, we compare absolute times required for training not the number of FL iterations as shown in Figure~\ref{fig:final_result}.

The proposed algorithm outperforms the two baselines i) because of the large training dataset it is capable of processing per unit time as our algorithm prefers computationally powerful devices as long as they are not used too many times as indicated by Equation~(\ref{eq:usefulness}), and ii) because of higher utilization as clients minimize the slack time by adjusting the training dataset size while the baseline algorithms inevitably wait for other clients to finish.

Moreover, compared to MinCost algorithm, the proposed method outperforms it as our method explicitly balances the class distribution in the FL round, where MinCost algorithm mostly looks at simply whether or not certain class is represented.
The number of images from a class may be insufficient to ensure test accuracy of the class.
This shows that client selection and resulting collective distribution of data is of paramount concern when it comes to convergence of the model.
Compared to probPart algorithm, the performance is slightly better than MinCost algorithm, but a random selection of clients results do not guarantee every class being represented in each FL iteration, and catastrophic forgetting for the classes lowers the overall test accuracy.

\begin{table}[t]
\centering
\caption{Per-class fairness on the CIFAR-100 test set at $\alpha=0.3$ for each method's final trained model (single run): overall accuracy, tail-class accuracy (mean of the 10 lowest-accuracy classes), and the dispersion of the per-class accuracies (coefficient of variation and Gini coefficient; lower is more uniform).}
\label{tab:fairness}
\footnotesize
\begin{tabular}{|l|c|c|c|c|}
\hline
\textbf{Method} & \textbf{Overall} & \textbf{Tail-10} & \textbf{CV} & \textbf{Gini} \\
                & \textbf{acc.\ (\%)} & \textbf{acc.\ (\%)} & & \\ \hline \hline
Proposed (full) & \textbf{42.3} & \textbf{15.3} & \textbf{0.43} & \textbf{0.24} \\
probPart        & 38.0 & 6.8 & 0.49 & 0.28 \\
MinCost         & 31.2 & 3.5 & 0.62 & 0.35 \\ \hline
\end{tabular}
\end{table}

To characterize the results beyond the aggregate accuracy, we examine how evenly the accuracy is distributed across the 100 classes, since a model that sacrifices under-represented classes is undesirable even at a similar mean accuracy. We quantify this with two standard, complementary measures of dispersion: the coefficient of variation (CV), i.e., the standard deviation of the per-class accuracies divided by their mean, which reflects how \emph{consistent} the accuracy is across classes (lower is more uniform); and the Gini coefficient, which reflects how \emph{equally} the accuracy is shared among classes (0 denotes perfect equality, while larger values indicate that accuracy is concentrated in fewer classes). We additionally report the tail-class accuracy, i.e., the mean accuracy of the 10 lowest-performing classes, which are the ones most exposed to forgetting. As summarized in Table~\ref{tab:fairness}, the proposed method more than doubles the tail-class accuracy of probPart (15.3\% versus 6.8\%) and quadruples that of MinCost (3.5\%), while attaining both the lowest CV and the lowest Gini. Hence, explicitly balancing the collective class distribution in each FL round keeps the accuracy both consistent (low CV) and equitably distributed (low Gini) across classes, so that the gain of the proposed algorithm is concentrated on the under-represented classes that the baselines tend to forget.

\subsection{Ablation and Sensitivity Analysis of the Usefulness Metric}


To address the individual contribution of each component of the usefulness
metric (Eq.~\ref{eq:usefulness}), we conduct an ablation study on the weights
$w_1$, $w_2$, and $w_3$. Since the $w_1$ term is, by construction, a product of
two distinct factors, i.e., a preference for a larger trainable dataset size
$|TD_k|$ and a freshness penalty $exp(-n_{avg,k}/10)$ that discourages data
reuse, we further decouple these two sub-factors to isolate their individual
effect. Each variant disables exactly one component while keeping the rest
unchanged, and is evaluated under the same deadline-based, equal-time protocol
as in Fig.~\ref{fig:final_result}, for two degrees of non-iidness
($\alpha=0.3$ and a more extreme $\alpha=0.1$).

As shown in Fig.~\ref{fig:ablation}, the freshness penalty is the single most influential component: disabling it prevents the model from reaching a useful accuracy at both severities, plateauing well below every other variant, and the degradation deepens as the data becomes more skewed ($\alpha=0.1$). 
The remaining terms, i.e., the dataset-size factor, the class-balance term $w_2$, and the coverage term $w_3$, each contribute a smaller and complementary gain, so that the full metric attains the highest accuracy. 
The ordering of the variants is preserved across $\alpha=0.3$ and $\alpha=0.1$, indicating that no single weight dominates in a way that would make the metric fragile to its exact values, and that the design generalizes across the degree of non-iidness.



\section{Conclusions}\label{sec:conclusions}
The paper proposed a joint client and training data selection algorithm for edge FL fast convergence of FL models.
The work puts emphasis on estimating training times, which is largely impacted by the training dataset size, and proposes an algorithm, which balances training datasize, class distribution per FL iteration and data staleness.
The experimental results display that our joint client selection and training data allocation algorithm achieves better FL convergence compared to other client and training dataset selection schemes by addressing the heterogeneity in computing capacities and non-iid data more explicitly.

In future, we plan to bring the work closer to real world by considering more concrete interruption models for selected applications and investigating dynamic aspects of FL applications.

We also plan to extend the study of the usefulness metric beyond the per-term ablation presented here, by examining how different combinations of the weights $w_1$, $w_2$, and $w_3$ jointly affect the accuracy, and by evaluating the algorithm under different client-pool distributions and selection ratios, as the current experiments consider a single setting of 10 out of 50 clients.

\small
\section*{Acknowledgments}
This work was partly supported by Einstein Center Digital Future, by Turkish Ministry of Education, and by JSPS KAKENHI Grant Number 23K16858.

\bibliographystyle{IEEEtran}
\bibliography{references}

@inproceedings{sandler2018mobilenetv2,
  title={Mobilenetv2: Inverted residuals and linear bottlenecks},
  author={Sandler, Mark and Howard, Andrew and Zhu, Menglong and Zhmoginov, Andrey and Chen, Liang-Chieh},
  booktitle={Proceedings of the IEEE conference on computer vision and pattern recognition},
  pages={4510--4520},
  year={2018}
}

@inproceedings {lai:usenix21,
author = {Fan Lai and Xiangfeng Zhu and Harsha V. Madhyastha and Mosharaf Chowdhury},
title = {Oort: Efficient Federated Learning via Guided Participant Selection},
booktitle = {USENIX Symposium on Operating Systems Design and Implementation ({OSDI} 21)},
year = {2021},
isbn = {978-1-939133-22-9},
pages = {19--35},
url = {https://www.usenix.org/conference/osdi21/presentation/lai},
publisher = {{USENIX} Association},
month = jul
}

@ARTICLE{chen:iot24,
  author={Chen, Xiaobing and Zhou, Xiangwei and Zhang, Hongchao and Sun, Mingxuan and Vincent Poor, H.},
  journal={IEEE Internet of Things Journal}, 
  title={Client Selection for Wireless Federated Learning With Data and Latency Heterogeneity}, 
  year={2024},
  volume={11},
  number={19},
  pages={32183-32196},
  doi={10.1109/JIOT.2024.3425757}}

@article{krizhevsky2009learning,
  title={Learning multiple layers of features from tiny images},
  author={Krizhevsky, Alex and Hinton, Geoffrey and others},
  journal={},
  year={2009},
  publisher={Toronto, ON, Canada}
}

@inproceedings{mcmahan2017communication,
  title={Communication-efficient learning of deep networks from decentralized data},
  author={McMahan, Brendan and Moore, Eider and Ramage, Daniel and Hampson, Seth and y Arcas, Blaise Aguera},
  booktitle={Artificial intelligence and statistics},
  pages={1273--1282},
  year={2017},
  organization={PMLR}
}

@article{bonawitz2019towards,
  title={Towards federated learning at scale: System design},
  author={Bonawitz, Keith and Eichner, Hubert and Grieskamp, Wolfgang and Huba, Dzmitry and Ingerman, Alex and Ivanov, Vladimir and Kiddon, Chloe and Kone{\v{c}}n{\`y}, Jakub and Mazzocchi, Stefano and McMahan, Brendan and others},
  journal={Proceedings of machine learning and systems},
  volume={1},
  pages={374--388},
  year={2019}
}

@article{lim2020federated,
  title={Federated learning in mobile edge networks: A comprehensive survey},
  author={Lim, Wei Yang Bryan and Luong, Nguyen Cong and Hoang, Dinh Thai and Jiao, Yutao and Liang, Ying-Chang and Yang, Qiang and Niyato, Dusit and Miao, Chunyan},
  journal={IEEE Communications Surveys \& Tutorials},
  volume={22},
  number={3},
  pages={2031--2063},
  year={2020},
  publisher={IEEE}
}

@article{gu2021server,
  title={From server-based to client-based machine learning: A comprehensive survey},
  author={Gu, Renjie and Niu, Chaoyue and Wu, Fan and Chen, Guihai and Hu, Chun and Lyu, Chengfei and Wu, Zhihua},
  journal={ACM Computing Surveys (CSUR)},
  volume={54},
  number={1},
  pages={1--36},
  year={2021},
  publisher={ACM New York, NY, USA}
}

@article{xiao2023time,
  title={Time-sensitive learning for heterogeneous federated edge intelligence},
  author={Xiao, Yong and Zhang, Xiaohan and Li, Yingyu and Shi, Guangming and Krunz, Marwan and Nguyen, Diep N and Hoang, Dinh Thai},
  journal={IEEE Transactions on Mobile Computing},
  year={2023},
  publisher={IEEE}
}

@article{reisizadeh2022straggler,
  title={Straggler-resilient federated learning: Leveraging the interplay between statistical accuracy and system heterogeneity},
  author={Reisizadeh, Amirhossein and Tziotis, Isidoros and Hassani, Hamed and Mokhtari, Aryan and Pedarsani, Ramtin},
  journal={IEEE Journal on Selected Areas in Information Theory},
  volume={3},
  number={2},
  pages={197--205},
  year={2022},
  publisher={IEEE}
}

@article{zhao2021federated,
  title={Federated learning with heterogeneity-aware probabilistic synchronous parallel on edge},
  author={Zhao, Jianxin and Han, Rui and Yang, Yongkai and Catterall, Benjamin and Liu, Chi Harold and Chen, Lydia Y and Mortier, Richard and Crowcroft, Jon and Wang, Liang},
  journal={IEEE Transactions on Services Computing},
  volume={15},
  number={2},
  pages={614--626},
  year={2021},
  publisher={IEEE}
}

@INPROCEEDINGS{kim:iccad15,
  author={Kim, Yeseong and Parterna, Francesco and Tilak, Sameer and Rosing, Tajana S.},
  booktitle={IEEE/ACM International Conference on Computer-Aided Design (ICCAD)}, 
  title={Smartphone analysis and optimization based on user activity recognition}, 
  year={2015},
  volume={},
  number={},
  pages={605-612},
  doi={10.1109/ICCAD.2015.7372625}}

@inproceedings{li2019smartpc,
  title={SmartPC: Hierarchical pace control in real-time federated learning system},
  author={Li, Li and Xiong, Haoyi and Guo, Zhishan and Wang, Jun and Xu, Cheng-Zhong},
  booktitle={2019 IEEE Real-Time Systems Symposium (RTSS)},
  pages={406--418},
  year={2019},
  organization={IEEE}
}

@inproceedings{cui2022helcfl,
  title={HELCFL: High-efficiency and low-cost federated learning in heterogeneous mobile-edge computing},
  author={Cui, Yangguang and Cao, Kun and Zhou, Junlong and Wei, Tongquan},
  booktitle={2022 Design, Automation \& Test in Europe Conference \& Exhibition (DATE)},
  pages={1227--1232},
  year={2022},
  organization={IEEE}
}

@inproceedings{shi2020device,
  title={Device scheduling with fast convergence for wireless federated learning},
  author={Shi, Wenqi and Zhou, Sheng and Niu, Zhisheng},
  booktitle={ICC 2020-2020 IEEE International Conference on Communications (ICC)},
  pages={1--6},
  year={2020},
  organization={IEEE}
}

@article{xu2021online,
  title={Online client scheduling for fast federated learning},
  author={Xu, Bo and Xia, Wenchao and Zhang, Jun and Quek, Tony QS and Zhu, Hongbo},
  journal={IEEE Wireless Communications Letters},
  volume={10},
  number={7},
  pages={1434--1438},
  year={2021},
  publisher={IEEE}
}

@article{xia2020multi,
  title={Multi-armed bandit-based client scheduling for federated learning},
  author={Xia, Wenchao and Quek, Tony QS and Guo, Kun and Wen, Wanli and Yang, Howard H and Zhu, Hongbo},
  journal={IEEE Transactions on Wireless Communications},
  volume={19},
  number={11},
  pages={7108--7123},
  year={2020},
  publisher={IEEE}
}

@article{amiri2021convergence,
  title={Convergence of update aware device scheduling for federated learning at the wireless edge},
  author={Amiri, Mohammad Mohammadi and G{\"u}nd{\"u}z, Deniz and Kulkarni, Sanjeev R and Poor, H Vincent},
  journal={IEEE Transactions on Wireless Communications},
  volume={20},
  number={6},
  pages={3643--3658},
  year={2021},
  publisher={IEEE}
}

@inproceedings{wadu2020federated,
  title={Federated learning under channel uncertainty: Joint client scheduling and resource allocation},
  author={Wadu, Madhusanka Manimel and Samarakoon, Sumudu and Bennis, Mehdi},
  booktitle={2020 IEEE Wireless Communications and Networking Conference (WCNC)},
  pages={1--6},
  year={2020},
  organization={IEEE}
}

@article{wadu2021joint,
  title={Joint client scheduling and resource allocation under channel uncertainty in federated learning},
  author={Wadu, Madhusanka Manimel and Samarakoon, Sumudu and Bennis, Mehdi},
  journal={IEEE Transactions on Communications},
  volume={69},
  number={9},
  pages={5962--5974},
  year={2021},
  publisher={IEEE}
}

@article{pilla2023scheduling,
  title={Scheduling algorithms for federated learning with minimal energy consumption},
  author={Pilla, La{\'e}rcio Lima},
  journal={IEEE Transactions on Parallel and Distributed Systems},
  volume={34},
  number={4},
  pages={1215--1226},
  year={2023},
  publisher={IEEE}
}

@inproceedings{amiri2020update,
  title={Update aware device scheduling for federated learning at the wireless edge},
  author={Amiri, Mohammad Mohammadi and G{\"u}nd{\"u}z, Deniz and Kulkarni, Sanjeev R and Poor, H Vincent},
  booktitle={2020 IEEE International Symposium on Information Theory (ISIT)},
  pages={2598--2603},
  year={2020},
  organization={IEEE}
}

@article{wang2019adaptive,
  title={Adaptive federated learning in resource constrained edge computing systems},
  author={Wang, Shiqiang and Tuor, Tiffany and Salonidis, Theodoros and Leung, Kin K and Makaya, Christian and He, Ting and Chan, Kevin},
  journal={IEEE journal on selected areas in communications},
  volume={37},
  number={6},
  pages={1205--1221},
  year={2019},
  publisher={IEEE}
}

@inproceedings{wang2020optimize,
  title={Optimize scheduling of federated learning on battery-powered mobile devices},
  author={Wang, Cong and Wei, Xin and Zhou, Pengzhan},
  booktitle={2020 IEEE International Parallel and Distributed Processing Symposium (IPDPS)},
  pages={212--221},
  year={2020},
  organization={IEEE}
}

@article{wang2020towards,
  title={Towards efficient scheduling of federated mobile devices under computational and statistical heterogeneity},
  author={Wang, Cong and Yang, Yuanyuan and Zhou, Pengzhan},
  journal={IEEE Transactions on Parallel and Distributed Systems},
  volume={32},
  number={2},
  pages={394--410},
  year={2020},
  publisher={IEEE}
}

@inproceedings{pilla2021optimal,
  title={Optimal task assignment for heterogeneous federated learning devices},
  author={Pilla, La{\'e}rcio Lima},
  booktitle={2021 IEEE International Parallel and Distributed Processing Symposium (IPDPS)},
  pages={661--670},
  year={2021},
  organization={IEEE}
}

@article{ma2022state,
  title={A state-of-the-art survey on solving non-IID data in Federated Learning},
  author={Ma, Xiaodong and Zhu, Jia and Lin, Zhihao and Chen, Shanxuan and Qin, Yangjie},
  journal={Future Generation Computer Systems},
  volume={135},
  pages={244--258},
  year={2022},
  publisher={Elsevier}
}

@article{yu2021jointly,
  title={Jointly optimizing client selection and resource management in wireless federated learning for internet of things},
  author={Yu, Liangkun and Albelaihi, Rana and Sun, Xiang and Ansari, Nirwan and Devetsikiotis, Michael},
  journal={IEEE Internet of Things Journal},
  volume={9},
  number={6},
  pages={4385--4395},
  year={2021},
  publisher={IEEE}
}

@inproceedings{gao2021fedswap,
  title={FedSwap: a federated learning based 5G decentralized dynamic spectrum access system},
  author={Gao, Zhihui and Li, Ang and Gao, Yunfan and Li, Bing and Wang, Yu and Chen, Yiran},
  booktitle={2021 IEEE/ACM International Conference On Computer Aided Design (ICCAD)},
  pages={1--6},
  year={2021},
  organization={IEEE}
}

@article{hsu2019measuring,
  title={Measuring the effects of non-identical data distribution for federated visual classification},
  author={Hsu, Tzu-Ming Harry and Qi, Hang and Brown, Matthew},
  journal={arXiv preprint arXiv:1909.06335},
  year={2019}
}

\vspace{12pt}

\end{document}